\documentclass[10pt,twocolumn,letterpaper]{article}

 \usepackage[pagenumbers]{iccv} 

\usepackage{multirow}
\usepackage{graphicx}
\usepackage{amsmath}
\usepackage{amssymb}
\usepackage{booktabs}
\newcommand{\mb}[1]{\mathbf{#1}}
\newcommand{\bs}[1]{\boldsymbol{#1}}

\definecolor{iccvblue}{rgb}{0.21,0.49,0.74}
\usepackage[pagebackref,breaklinks,colorlinks,allcolors=iccvblue]{hyperref}

\def\paperID{11608} 
\def\confName{ICCV}
\def\confYear{2025}

\title{High-Resolution Spatiotemporal Modeling with Global-Local State Space Models for Video-Based Human Pose Estimation}

\author{
    Runyang Feng\textsuperscript{\rm 1,2},
    Hyung Jin Chang\textsuperscript{\rm 3},
    Tze Ho Elden Tse\textsuperscript{\rm 4},
    Boeun Kim\textsuperscript{\rm 5},
    Yi Chang\textsuperscript{\rm 1,2},
    Yixing Gao\textsuperscript{\rm 1,2}\thanks{Corresponding Author (gaoyixing@jlu.edu.cn)}
    \\    
    \textsuperscript{1} School of Artificial Intelligence, Jilin University,\\
    \textsuperscript{2} Engineering Research Center of Knowledge-Driven Human-Machine Intelligence, \\Ministry of Education, China,
    \textsuperscript{3}School of Computer Science, University of Birmingham,\\
     \textsuperscript{4}National University of Singapore, \textsuperscript{5}Dankook University
}

\begin{document}
\maketitle
\begin{abstract}
Modeling high-resolution spatiotemporal representations, including both global dynamic contexts (e.g., holistic human motion tendencies) and local motion details (e.g., high-frequency changes of keypoints), is essential for video-based human pose estimation (VHPE). Current state-of-the-art methods typically unify spatiotemporal learning within a single type of modeling structure (convolution or attention-based blocks), which inherently have difficulties in balancing global and local dynamic modeling and may bias the network to one of them, leading to suboptimal performance. Moreover, existing VHPE models suffer from quadratic complexity when capturing global dependencies, limiting their applicability especially for high-resolution sequences. Recently, the state space models (known as Mamba) have demonstrated significant potential in modeling long-range contexts with linear complexity; however, they are restricted to 1D sequential data. In this paper, we present a novel framework that extends Mamba from two aspects to separately learn global and local high-resolution spatiotemporal representations for VHPE. Specifically, we first propose a Global Spatiotemporal Mamba, which performs 6D selective space-time scan and spatial- and temporal-modulated scan merging to efficiently extract global representations from high-resolution sequences. We further introduce a windowed space-time scan-based Local Refinement Mamba to enhance the high-frequency details of localized keypoint motions. Extensive experiments on four benchmark datasets demonstrate that the proposed model outperforms state-of-the-art VHPE approaches while achieving better computational trade-offs.
\end{abstract}    
\vspace{-2em}
\section{Introduction}
\label{sec:intro}
Human pose estimation is a fundamental task in computer vision that has attracted increasing attention in recent years. The objective is to detect and localize anatomical human keypoints, such as elbows and wrists, from still images or video sequences. It finds enormous applications in diverse realistic scenes including human behavior understanding, augmented reality, and surveillance tracking~\cite{sun2019deep, feng2023mutual}.

\begin{figure}
\begin{center}
\includegraphics[width=.93\linewidth]{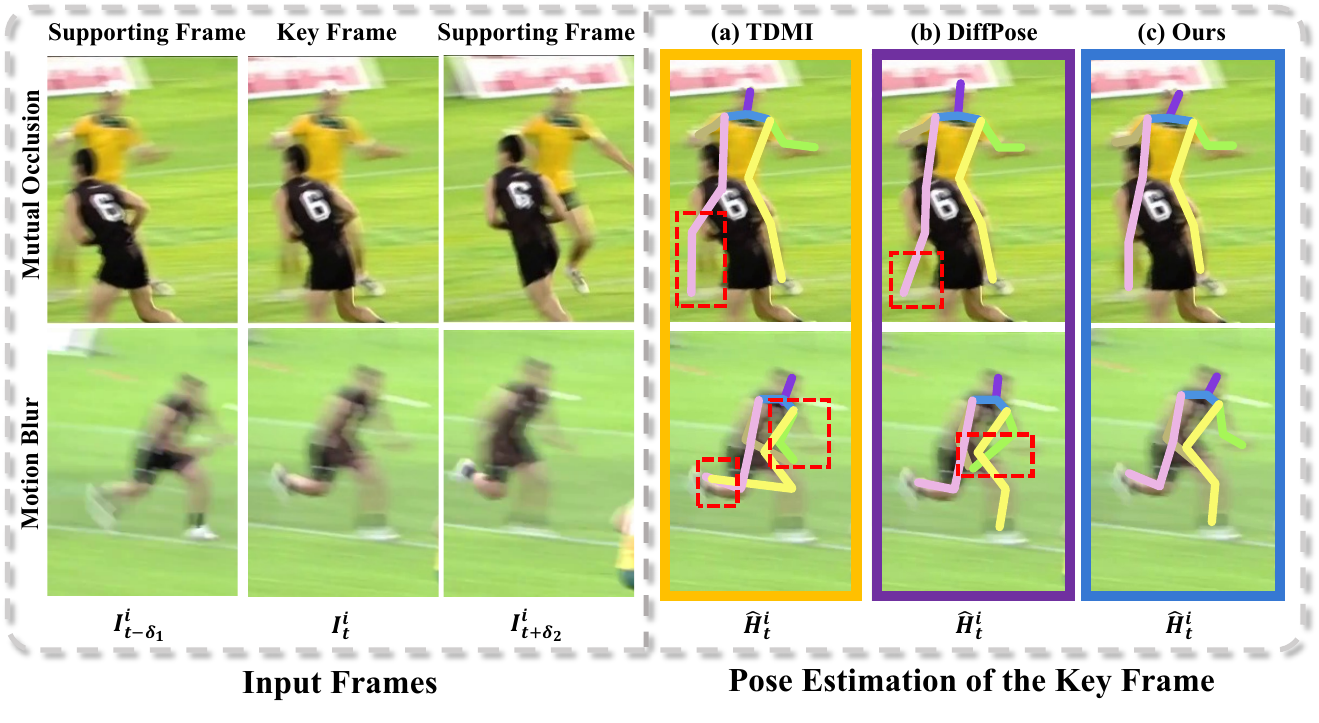}
\end{center}
\vspace{-1.2em}
\caption{State-of-the-art methods such as (a) TDMI~\cite{feng2023mutual} and (b) DiffPose~\cite{feng2023diffpose} focus either on global or local spatiotemporal contexts, which may fail for occlusion or blur cases. Our method (c) fully exploits both global and local high-resolution spatiotemporal representations, delivering more robust results.} \label{fig:paradigm}
\vspace{-1.5em}
\end{figure}

Accurately estimating human poses from videos requires dense spatiotemporal analysis, which significantly benefits from global and local high-resolution representations~\cite{wang2020deep, yu2021lite}. The former typically characterizes holistic human motion patterns and contexts, while the later captures detailed high-frequency variations of local keypoints. With the surge in deep learning, numerous VHPE approaches using a single type of modeling structure such as convolutions~\cite{he2016deep} or Transformers~\cite{dosovitskiy2020image} have been designed.

The CNN-based methods~\cite{bertasius2019learning, liu2022temporal, feng2023mutual} usually design convolutional networks to integrate spatial and temporal information derived from HRNet (an off-the-shelf model for extracting high-resolution image features).
For instance, \cite{feng2023mutual} computes feature differences among frames to capture motion clues, and aggregates high-resolution appearance and motion features using convolutions to estimate pose heatmaps. \cite{liu2022temporal} employs convolutions to align multiple supporting frames to the keyframe and fuses all aligned feature maps for pose estimation. However, the fixed receptive fields inherent in convolutions constrain the global inference capability of these approaches, which may result in large prediction deviations for degraded body parts in challenging cases. As illustrated in Fig.~\ref{fig:paradigm} (a), TDMI~\cite{feng2023mutual} produces inaccurate estimations for right leg in mutual occlusion scenes (or left arm for blur cases). In contrast, Transformer-based methods~\cite{feng2023diffpose, he2024video} adopt a self-attention mechanism, allowing them to capture global dependencies of the input sequence. \cite{feng2023diffpose} concatenates the features of each frame and employs plain Vision Transformers to obtain global spatiotemporal representations. Nevertheless, the attention-based models often suffer from inferior local high-frequency details, and tend to yield inaccurate detections for localized ambiguous joints (right ankle in top of Fig.~\ref{fig:paradigm} (b) or left wrist in bottom). Moreover, the computation of self-attention involves quadratic complexity with respect to input tokens. Directly applying self-attention to high-resolution sequences (\emph{e.g.}, $1/4*T$) would result in excessive computation and memory overheads (Table~\ref{run}).

Recently, state space models (SSMs) have gained significant attention for their strengths in capturing long-range dependencies~\cite{gu2021efficiently, zhu2024vision}. Notably, Mamba~\cite{gu2023mamba} incorporates parallelized selective scan and a hardware-aware algorithm, achieving remarkable performance in long language modeling with linear complexity. 
Despite these merits, Mamba's core operator, the \emph{vanilla} selective scan, is specially designed for 1D sequential data. 
This presents substantial challenges when adapting to the spatiotemporal information in videos.
Recent variants~\cite{park2024videomamba, li2024videomamba} attempt to extend Mamba to video processing (\emph{e.g.}, high-level video understanding) via frame-by-frame bidirectional scanning. They simply flatten the spatial tokens of each frame within the sequence to model global dependencies. 
 However, such scanning schemes focus on sequential spatial processing and do not take into account of insights from other scanning directions, which is detrimental to the dense analysis of high-resolution spatiotemporal contexts in VHPE. For instance, they elongate the distance between temporally adjacent tokens, leading to insufficient capture of temporal-wise pixel dynamics.
 Moreover, these methods lack specific designs to guide the Mamba to learn fine-grained local details from sequences. Directly applying them to VHPE produces inferior performance.

Inspired by the preceding analysis, we design a decoupled framework based on pure \underline{\textbf{Mamba}} to explore \underline{\textbf{G}}lobal and \underline{\textbf{L}}ocal high-resolution \underline{\textbf{S}}patiotemporal representations for VHPE (GLSMamba). The proposed GLSMamba extends Mamba in two aspects: \textbf{(i)} A Global Spatiotemporal Mamba (GSM) is designed for holistic contextual sequence modeling at high resolutions. Specifically, GSM engages a \emph{6D selective Space-Time Scan (STS6D)} mechanism, which traverses along six tailored spatiotemporal scanning routes to fully resolve the high-resolution feature sequences from a global perspective. Then, GSM adaptively aggregates the scanning knowledge from different routes via a \emph{Spatial- and Temporal-Modulated scan Merging (STMM)} strategy, thereby bridging the gap between 1D selective scan and high-resolution sequences. \textbf{(ii)} A Local Refinement Mamba (LRM) is further introduced to enhance the high-frequency details of local motion representations. LRM performs \emph{frame-wise selective scan within windowed patch cubes}, processing localized pixels inside the same semantic 3D tubelet compactly together to effectively capture local spatiotemporal dependencies. This module significantly enhances fine-grained motion details while preserving sequence-size receptive field. Thanks to the Mamba-based decoupled structure design, our approach delivers more reliable high-resolution spatiotemporal representations that are globally consistent and locally enriched, and possesses better computational trade-offs.

From extensive evaluations on four widely-used benchmark datasets (PoseTrack2017, PoseTrack2018, PoseTrack21, and Sub-JHMDB), we show that GLSMamba surpasses state-of-the-art VHPE methods. We also provide ablation analysis on the effectiveness of each proposed component and design choice.

The key contributions of this work can be summarized as: (i) We propose to decouple the modeling of global dynamic contexts and local motion details for video-based human pose estimation. (ii) We present GLSMamba, the first pure Mamba-based framework for VHPE. GLSMamba extends the \emph{vanilla} state space model in two ways, forming GSM and LRM to learn global and local high-resolution spatiotemporal representations, respectively. (iii) We demonstrate that GLSMamba achieves competitive state-of-the-art performance with fewer parameters on four benchmark datasets: PoseTrack2017, PoseTrack2018, PoseTrack21, and Sub-JHMDB.

\section{Related Work}
\label{sec:related}

\begin{figure*}
\begin{center}
\includegraphics[width=.85\linewidth]{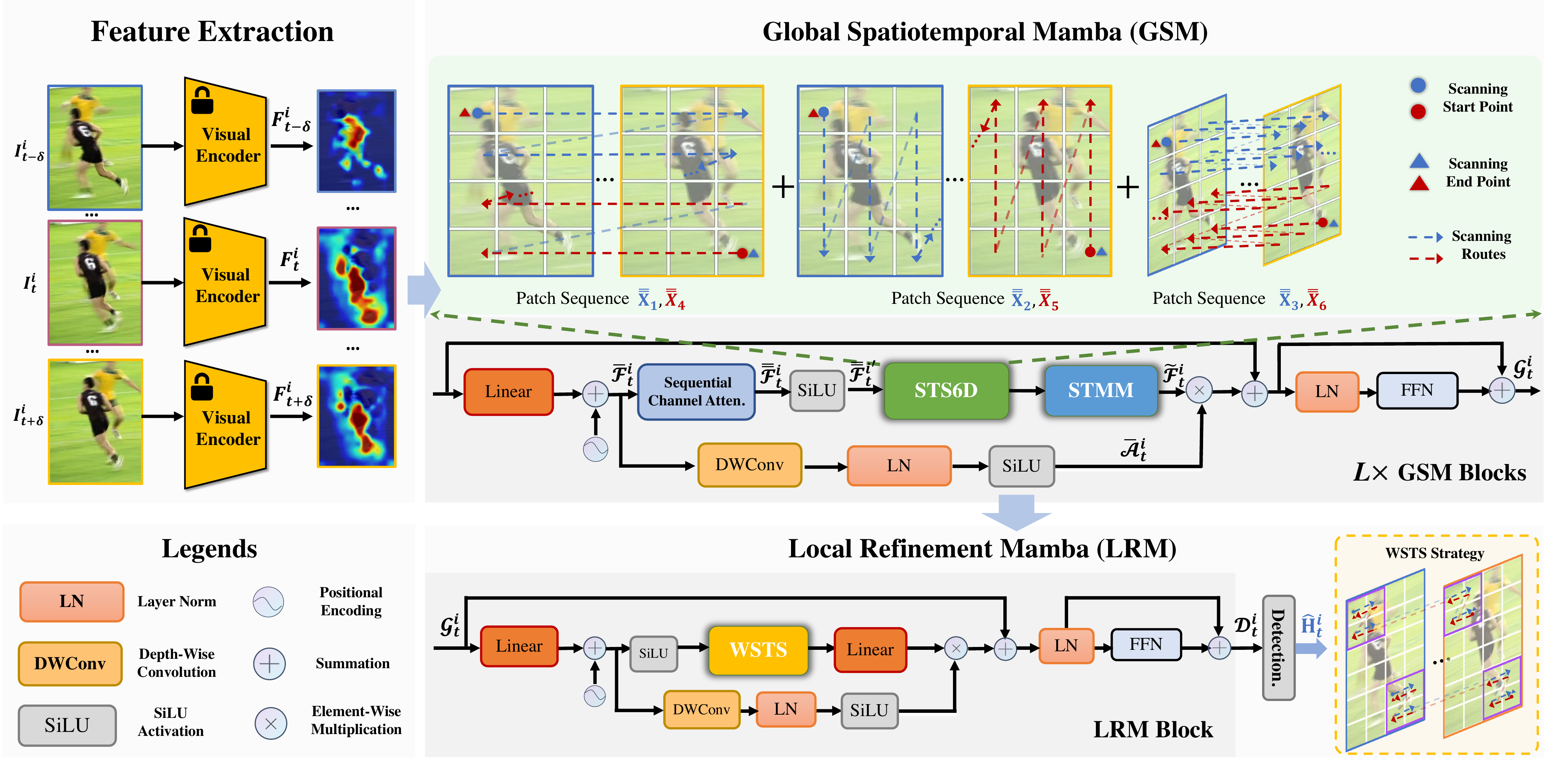}
\end{center}
\vspace{-1em}
\caption{Overall pipeline of the proposed framework. Given an input sequence, we first extract high-resolution spatial features for each frame using a visual encoder. Then, these features are processed successively by GSM and LRM for global spatiotemporal modeling and local detail enhancement. Finally, a detection head is employed to yield the pose heatmap estimations.}
\vspace{-1.3em}
\label{fig:pipeline}
\end{figure*}

\textbf{Human pose estimation in images.}\quad
Estimating human joint locations from still images has been extensively studied which generally falls into two paradigms: bottom-up and top-down. \emph{Bottom-up} approaches~\cite{cao2017realtime, kocabas2018multiposenet, kreiss2019pifpaf, cheng2020higherhrnet} first detect individual body parts and then group them into an entire human skeleton. 
The main variation among these methods lies in the grouping algorithms, such as Part Affinity Field in \cite{cao2017realtime} and Associative Embedding in \cite{newell2017associative}. 
Conversely, \emph{top-down} approaches~\cite{sun2019deep, xiao2018simple, li2021tokenpose, xu2022vitpose} first extract human bounding boxes using an object detector, and then design models to estimate human poses within each bounding box region. \cite{sun2019deep} presents a high-resolution convolutional network that retains high-resolution feature maps throughout all stages and performs repeated multi-scale fusion to obtain rich human body information. \cite{xu2022vitpose} leverages cascaded plain vision transformers to learn generalizable image representations, achieving superior performance on multiple benchmarks.

\noindent
\textbf{Human pose estimation in videos.}\quad
Directly applying existing image-based pose estimation models to videos often yields suboptimal results because they fail to capture the temporal dynamics among frames. To incorporate spatiotemporal contexts, several state-of-the-art approaches~\cite{bertasius2019learning, liu2021deep, liu2022temporal, feng2023mutual} integrate HRNet \cite{sun2019deep} as the backbone network, and adopt a convolutional architecture to aggregate high-resolution spatial and temporal feature representations. \cite{bertasius2019learning, liu2021deep} compute joint motion offsets between frames and employ the motion information to guide accurate pose heatmap resampling. 
 \cite{feng2023mutual} introduces temporal feature differences as the motion clues, and employs convolutional blocks to aggregate appearance and motion features. A primary drawback of these methods lies in the restricted global spatiotemporal perception ability due to the limited receptive field, which may hinder their performance. 
   Another line of work \cite{feng2023diffpose, he2024video} considers Transformers (self-attentions) for global spatiotemporal modeling. \cite{feng2023diffpose} extracts feature tokens for each frame, and then employs Vision Transformers to capture the global dependencies of the token sequence. However, these methods often neglect valuable high-frequency details of local keypoint motions, and incur quadratic computational complexity that is detrimental to high-resolution modeling.
   Different from the above methods, we aim to introduce a novel decoupled architecture to fully learn both global and local high-resolution spatiotemporal contexts for VHPE while maintaining an acceptable computational load.

\noindent  
\textbf{State space models.}\quad
 State space models (SSMs) are a type of foundation models and have recently demonstrated great potential in capturing long range dependencies through HiPPO matrix initialization~\cite{gu2021combining, gu2020hippo}. To facilitate the practical applicability of SSMs, \cite{gu2021efficiently} proposes the structured SSM model (S4) which imposes a diagonalization structure on the parameter matrix, significantly reducing the computational overhead. The promising results from S4 have inspired the emergence of numerous SSM-based architectures. For example, S5~\cite{smith2022simplified} proposes a multi-input and multi-output SSM, GSS~\cite{mehta2022long} integrates a gated mechanism, and the recent advancement Mamba~\cite{gu2023mamba} introduces context-based reasoning and parallelized selective scanning. Due to the exceptional performance in long sequence modeling with linear complexity, Mamba has become a compelling alternative to Transformers, finding extensive applications in diverse fields ranging from language and audio~\cite{gu2023mamba} to vision tasks~\cite{liu2024vmambavisualstatespace, zhu2024vision, park2024videomamba}.

 Despite several visual Mamba variants for action recognition~\cite{park2024videomamba, li2024videomamba}, these models have difficulties in adequately processing dense high-resolution sequence contexts. They also lack specific designs to capture local spatiotemporal details.  
 In contrast, we purposefully extend Mamba for VHPE from two aspects, with a focus on learning global and local high-resolution spatiotemporal representations.

\vspace{-.3em}
\section{Our Approach}
\label{sec:approach}

\textbf{Problem formulation.}\quad Our work follows a top-down paradigm in which a human detector is first used to obtain the human bounding boxes in a frame $I_t$. Then, each of the bounding boxes is enlarged by $125\%$ to crop the same individual $i$ across a frame sequence $\bs{\mathcal{I}_t^i} = \left\langle I_{t-\delta}^i, ..., I_t^i, ..., I_{t+\delta}^i \right\rangle$, where $\delta$ denotes the temporal span. Given $\bs{\mathcal{I}_t^i}$, we seek to explore the spatiotemporal clues to foster the pose estimation in the current frame $I_t^i$  .

\noindent
\textbf{Method overview.}\quad
The overall pipeline of our proposed GLSMamba framework is shown in Fig.~\ref{fig:pipeline}. Our objective is to extend Mamba to model global-local high-resolution spatiotemporal contexts effectively.
There are two key components: Global Spatiotemporal Mamba (GSM) and Local Refinement Mamba (LRM). Specifically, we first extract high-resolution features for each frame $\bs{\mathcal{F}_t^i} = \left\langle F_{t-\delta}^i, ..., F_t^i, ..., F_{t+\delta}^i \right\rangle$ using a visual encoder. Then, these features are fed into GSM for global spatiotemporal modeling.
The resulting tensor $\bs{\mathcal{G}_t^i}$ is passed to LRM to enhance the local spatiotemporal details and yield $\bs{\mathcal{D}_t^i}$. Finally, a detection head is used to estimate the pose heatmap $\hat{\mb{H}}_t^i$. 
In the following, we present the preliminaries of SSMs (Sec.~\ref{sec:Preliminaries}), and detail the architectures of each component including GSM (Sec.~\ref{sec:GSM}) and LRM (Sec.~\ref{sec:LDR}).

\subsection{Preliminaries}
\label{sec:Preliminaries}

\textbf{State space models.}  SSMs are inspired by the continuous system that maps a 1D input signal to output response $x(n) \in \mathbb{R} \mapsto y(n) \in \mathbb{R}$ via a hidden state $\mathbf{h}(n) \in \mathbb{R}^N$. Formally, SSMs can be expressed as the following ordinary differential equations:
\begin{equation}\label{eq:ssm}
	\begin{aligned}
		\mathbf{h}'(n) &= \mb{A}\mb{h}(n) + \mb{B}x(n), \\
		y(n) &= \mb{Ch}(n) + Dx(n),
	\end{aligned}
\end{equation}
where $\mb{A} \in \mathbb{R}^{N\times N}$ denotes the evolution parameter, and $\mb{B} \in \mathbb{R}^{N\times 1}$, $\mb{C} \in \mathbb{R}^{1\times N}$ are projection matrixes. The parameter $D \in \mathbb{R}^{1}$ can be ignored as a residual connection.

To apply in the deep learning context, the above continuous system has to be discretized. 
The commonly used technique for discretizing SSMs, known as zero-order hold (ZOH), incorporates a step size $\bs{\Delta}$ to convert the continuous parameters $\mb{A}$ and $\mb{B}$ into their discrete counterparts, $\overline{\mb{A}}$ and $\overline{\mb{B}}$, respectively. This can be defined as:
\begin{equation} \label{eq:dis}
	\begin{aligned}
		\overline{\mb{A}} &= \exp \left(\bs{\Delta}\mb{A} \right), \\
		\overline{\mb{B}} &= \left(\bs{\Delta}\mb{A}\right)^{-1} \left(\exp \left(\bs{\Delta}\mb{A}\right)-\mb{I}\right) \cdot \bs{\Delta}\mb{B}.
	\end{aligned}
\end{equation}
Consequently, the continuous-time SSMs in Eqs.~\ref{eq:ssm} can be rewritten as:
\begin{equation}\label{DSSM}
	\begin{aligned}
		\mb{h}_n &= \overline{\mb{A}}\mb{h}_{n-1} + \overline{\mb{B}} x_n, \\
		y_n &= \mb{C}\mb{h}_n,
	\end{aligned}
\end{equation}
 which can be efficiently computed via global convolutions.

\noindent
\textbf{Selective SSM.}\quad 
A key property of the aforementioned SSM models is linear time invariance (LTI), implying that the parameters $\left(\mb{A}, \mb{B}, \mb{C}, \bs{\Delta}\right)$ remain constant and independent of the input across different time steps. To overcome this limitation, Mamba~\cite{gu2023mamba} introduces a selective scan mechanism (S6) as the core operator. Unlike LTI SSMs, Mamba dynamically generates model parameters based on the input, enabling context-based reasoning with linear complexity. Given these advantages, our work also adopts the S6 operation as a foundation.

\subsection{Global Spatiotemporal Mamba}
\label{sec:GSM}
As illustrated in Fig.~\ref{fig:pipeline}, we introduce the Global Spatiotemporal Mamba (GSM) to learn high-resolution spatiotemporal representations from a global perspective. To achieve this, we first construct the high-resolution feature sequence $\bs{\mathcal{F}_t^i}$ for the input clip $\bs{\mathcal{I}_t^i}$. The sequence $\bs{\mathcal{F}_t^i}$ is then passed through cascaded GSM blocks to produce the spatiotemporal features $\bs{\mathcal{G}_t^i}$.

\noindent
\textbf{High-resolution feature sequence extraction.}\quad Given $\bs{\mathcal{I}_t^i} = \left\langle I_{t-\delta}^i, ..., I_t^i \in \mathbb{R}^{\mathtt{C}\times \mathtt{H}\times \mathtt{W}}, ..., I_{t+\delta}^i \right\rangle$, a visual encoder pretrained on COCO is first leveraged to extract the features of each frame $\bs{\mathcal{F}_t^i} = \left\langle F_{t-\delta}^i, ..., F_t^i, ..., F_{t+\delta}^i \right\rangle$, where $(\mathtt{H}, \mathtt{W})$ indicates the image size and $\mathtt{C}$ is the number of channels. To ensure the high spatial resolution, we utilize ViTPose~\cite{xu2022vitpose} to extract image features followed by deconvolution structures for spatially upsampling by $4\times$. Note that the parameters of the visual encoder are frozen during the model optimization.

\noindent
\textbf{GSM block.}\quad
After obtaining the feature sequence $\bs{\mathcal{F}_t^i}$, we design the GSM block with the novel 6D selective Space-Time Scan (STS6D) and Spatial- and Temporal-Modulated scan Merging (STMM) mechanisms to model holistic spatiotemporal contexts. The core parts, STS6D and STMM, will be detailed in the following section. Specifically, the feature of each frame within $\bs{\mathcal{F}_t^i}$ is first linearly projected to a tensor with size $\mathtt{D}$, and combined with a sine-cosine spatial embedding $\mb{E}_{spa}$~\cite{vaswani2017attention} as well as a learnable temporal embedding  $\mb{E}_{tem}$ to preserve the spatial and temporal information, deriving $\bs{\bar{\mathcal{F}}_t^{i}} = \left\langle \bar{F}_{t-\delta}^i, ..., \bar{F}_t^i\in\mathbb{R}^{\mathtt{D}\times\mathtt{h} \times \mathtt{w}}, ..., \bar{F}_{t+\delta}^i \right\rangle$:
\begin{equation}\vspace{-1mm}
	\begin{aligned}
		\bs{\bar{\mathcal{F}}_t^i} = \mathrm{Linear}\left( \bs{\mathcal{F}_t^i} \right) + \mb{E}_{spa} + \mb{E}_{tem},
	\end{aligned}
\end{equation}
where $\mathtt{h} = \frac{1}{4}\mathtt{H}$ and $\mathtt{w} = \frac{1}{4}\mathtt{W}$. Subsequently, $\bs{\bar{\mathcal{F}}_t^i}$ is processed through two separate streams: 

\emph{(1) Main Stream:} To facilitate the global sequence modeling in STS6D and STMM, we first introduce a Sequential Channel Attention which adaptively activates significant spatiotemporal information within $\bs{\bar{\mathcal{F}}_t^i}$ at the channel level. Specifically, we concatenate the feature sequence and squeeze the global spatiotemporal information into sequential (frame-wise) channel descriptors via a global average pooling (GAP) layer. Next, several MLPs are leveraged to model channel interactions both spatially (intra-frame) and temporally (inter-frame), followed by a sigmoid function to obtain the sequential attention weights. The attention matrix ($\langle {M}_{t-\delta}^i, ..., {M}_t^i\in\mathbb{R}^{\mathtt{D}\times 1}, ..., {M}_{t+\delta}^i \rangle$) is used to rescale the input sequence $\bs{\bar{\mathcal{F}}_t^i}$ to obtain the modulated version $\bs{\bar{\bar{\mathcal{F}}}_t^i}$. The above process is formulated as:
\begin{equation}
	\begin{aligned}
		\bs{\bar{\bar{\mathcal{F}}}_t^i} = \sigma\left(\mathrm{MLPs}\left(\mathrm{GAP}\left(\bs{\bar{\mathcal{F}}_t^i}\right)\right)\right) \otimes \bs{\bar{\mathcal{F}}_t^i}.
	\end{aligned}
\end{equation}
We then normalize and use SiLU~\cite{shazeer2020glu} to transform $\bs{\bar{\bar{\mathcal{F}}}_t^i}$, and feed the resulting tensor $\bs{\bar{\bar{\mathcal{F}}}_t^{i'}}$ into STS6D and STMM to model global spatiotemporal dependencies and output $\bs{\tilde{\mathcal{F}}_t^i}$.

\emph{(2) Another stream} serves as a gated attention to further control the raw feature element propagation meticulously, which passes $\bs{\bar{\mathcal{F}}_t^i}$ into a depth-wise convolution, followed by a LayerNorm and a SiLU activation to yield $\bs{\bar{\mathcal{A}}_t^i}$. 

Finally, the resulting features of these two branches are merged via multiplication, and fed into a feedforward neural network (FFN) to obtain the global spatiotemporal representations $\bs{\mathcal{G}_t^i}$. In practice, we stack $L=4$ GSM blocks for progressive information processing.

\noindent
\textbf{STS6D and STMM.}\quad
Although the \emph{vanilla} selective scan in S6 enjoys various advantages such as global modeling, context-aware inference, and linear complexity, it is designed for 1D sequential data that differs substantially from the video modality. 
To address this challenge, we design the 6D selective Space-Time Scan (STS6D) as well as the Spatial- and Temporal-Modulated scan Merging (STMM) modules, which adapt S6 to high-resolution spatiotemporal modeling while maintaining its strengths.

\begin{figure}
\begin{center}
\includegraphics[width=.93\linewidth]{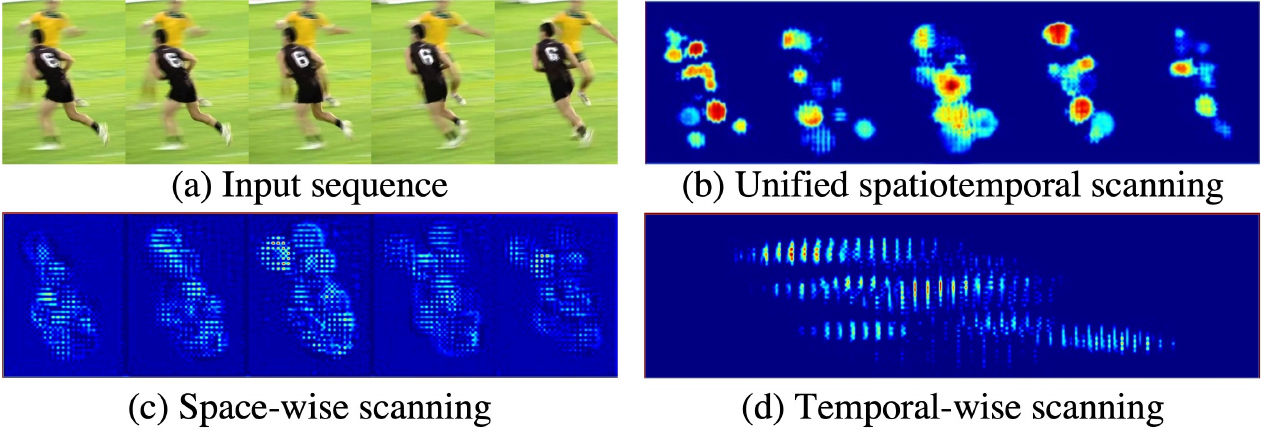}
\end{center}
\vspace{-1.4em}
\caption{Visualizations of activation maps of STS6D.}
\vspace{-1.8em}
\label{fig:scan}
\end{figure}

As illustrated in Fig.~\ref{fig:pipeline}, we first flatten $\bs{\bar{\bar{\mathcal{F}}}_t^{i'}}$ along six tailored space-time routes to obtain 1D patch sequences $\{{\bar{\bar{\mb{x}}}_k}\}_{k=1,2,...,6}$. Specifically, we stack the features of each frame within $\bs{\bar{\bar{\mathcal{F}}}_t^{i'}}$ to form an image-like panoramic spatiotemporal representation, and traverse it horizontally and vertically to yield ${{\bar{\bar{\mb{x}}}_1}}$ and ${{\bar{\bar{\mb{x}}}_2}}$. We further perform pixel traversal along the depth (time) dimension across frames to attain ${{\bar{\bar{\mb{x}}}_3}}$. Reversing $\{{\bar{\bar{\mb{x}}}_k}\}_{k=1,2,3}$ produces the complete six-way scanning sequences. 
 Then, each sequence is processed by a separate S6 block to capture the corresponding global dependencies: 
\begin{equation} 
	\begin{aligned}
		&\mb{B}_k = {f}_B\left({\bar{\bar{\mb{x}}}_k}\right), 
		 \mb{C}_k = {f}_C\left({\bar{\bar{\mb{x}}}_k}\right), 
		  \bs{\Delta}_k = {f}_{\Delta}\left({\bar{\bar{\mb{x}}}_k}\right),\\
		  &\overline{\mb{A}_k}, \overline{\mb{B}_k} = \mathrm{Dis}\left(\bs{\Delta}_k, \mb{A}_k, \mb{B}_k\right),\\
		  &\tilde{\mb{y}}_k = \mathrm{SSM}(\overline{\mb{A}_k}, \overline{\mb{B}_k}, \mb{C}_k)\left(\bs{\bar{\bar{\mb{x}}}_k}\right),
	\end{aligned}
\end{equation}
where $k \in \{1,2,...,6\}$, $(f_B, f_C, f_\Delta)$ refers to independent linear projections to generate parameters $(\mb{B}, \mb{C}, \bs{\Delta})$, and $\mb{A}$ is a learnable matrix with random initialization. The symbol  $\mathrm{Dis}(\cdot)$ denotes the discretization progress in Eqs.~\ref{eq:dis}, and $\mathrm{SSM}(\cdot)$ indicates the computations of state space model in Eqs.~\ref{DSSM}. \emph{Intuitively, the selective scanning of different routes can characterize a video clip from diverse views.} For instance, the unified scanning  $\{\tilde{\mb{y}}_1, \tilde{\mb{y}}_4\}$ captures high-level spatiotemporal representations of salient global dynamic contexts, as shown in Fig.~\ref{fig:scan} (b). In contrast, the space-wise scanning $\{\tilde{\mb{y}}_2, \tilde{\mb{y}}_5\}$ provides complete human spatial contexts of each frame, while the time-wise scanning $\{\tilde{\mb{y}}_3, \tilde{\mb{y}}_6\}$ approximates the dense motion tendencies of human body. 

Subsequently, given the processed feature sequences $\{\tilde{\mb{y}}_k\}_{k=1,2,...,6}$ with different semantics, a STMM mechanism is further proposed to adaptively aggregate them and yield $\bs{\tilde{\mathcal{F}}_t^i}$. To be specific, we first invert the backward scan sequences, and merge features belonging to the same type of scanning via an addition operation:
\begin{equation}
	\begin{aligned}
		  \tilde{\mb{y}}_u = \tilde{\mb{y}}_1 + \mathrm{Iv}(\tilde{\mb{y}}_4), 
		  \tilde{\mb{y}}_s = \tilde{\mb{y}}_2 + \mathrm{Iv}(\tilde{\mb{y}}_5), 
		  \tilde{\mb{y}}_t = \tilde{\mb{y}}_3 + \mathrm{Iv}(\tilde{\mb{y}}_6),
	\end{aligned}
\end{equation}
where $\mathrm{Iv}(\cdot)$ is the inverse transformation, tensors $\tilde{\mb{y}}_u$, $\tilde{\mb{y}}_s$, and $\tilde{\mb{y}}_t$ denote the high-level (unified), space-wise, and time-wise representations, respectively.
Then, we perform spatial-modulated and and temporal-modulated feature compensation to progressively update the high-level spatiotemporal features $\tilde{\mb{y}}_u$. 
Given $\tilde{\mb{y}}_u$ and $\tilde{\mb{y}}_s$, we reshape them to 2D sequences and concatenate them in the channel dimension. We then leverage convolutions to adaptively generate the kernel sampling offsets $\bs{\mathcal{O}}_{u;s}$ for $\tilde{\mb{y}}_u$, facilitating the learning of spatial compensation from $\tilde{\mb{y}}_s$. We also estimate the modulated scalars $\bs{\mathcal{W}}_{u;s}$ to control the sampling intensity. Finally, we conduct feature modulation via a deformable convolution (DCN~\cite{zhu2019deformable}) to update $\tilde{\mb{y}}_u$ as:
\vspace{-.5em}
\begin{equation}\vspace{-.5em}
	\begin{aligned}
		  \tilde{\mb{y}}_{u;s} = \tilde{\mb{y}}_u + \mathrm{DCN}\left(\tilde{\mb{y}}_u, \bs{\mathcal{O}}_{u;s}, \bs{\mathcal{W}}_{u;s}\right).
	\end{aligned}
\end{equation}
Similarly, the temporal-guided feature modulation is further performed over $\tilde{\mb{y}}_{u;s}$ and $\tilde{\mb{y}}_t$ for dense temporal compensation, obtaining $\bs{\tilde{\mathcal{F}}_t^i}$.

By thoroughly traversing the whole space-time domain and adaptively aggregating the multi-source scanning knowledge, GSM empowers each pixel to gather insights from all others across multiple directions. This facilitates the comprehensive processing and resolving of high-resolution sequences from a global perspective.

\subsection{Local Refinement Mamba}
\label{sec:LDR}
The tensor $\bs{\mathcal{G}_t^i}$ derived from GSM attends to the global understanding of human motion patterns, yet lacks rich local details of keypoints. To further enhance the fine-grained local spatiotemporal representations, we propose the Local Refinement Mamba (LRM) using a Windowed Space-Time Scan (WSTS) strategy. WSTS processes local pixels within a windowed 3D tubelet closely together to capture local spatiotemporal dependencies.

Specifically, WSTS first splits the input feature sequences into a series of local windowed temporal tubes (\emph{e.g.} $8 \times 6 \times T$). Then, a frame-wise selective scan is performed within each localized 3D tubelet. Concretely, each windowed feature tubelet is unrolled frame-by-frame in both forward and reverse directions, and then fed into a S6 block separately. We reshape the processed features and sum them to obtain locally enhanced spatiotemporal representations. Notably, WSTS leverages a non-overlapping window partition scheme to maintain the computational efficiency.

In our implementation, we remove the Sequential Channel Attention from the Global Spatiotemporal Mamba (GSM) block, and replace core operators \emph{i.e.} STS6D and STMM with the proposed WSTS strategy to construct the LRM block. We employ $2$ cascade LRM blocks to process the input tensor $\bs{\mathcal{G}_t^i}$, and obtain refined representations $\bs{\mathcal{D}_t^i}$ with abundant local details. 
 Finally, we aggregate the features of each frame within $\bs{\mathcal{D}_t^i}$ via an element-wise addition, and feed the resulting tensor into a detection head ($3 \times 3$ convolution) to yield the predicted pose heatmaps $\hat{\mb{H}}_t^i$.

\subsection{Loss Function}
\label{train}
 We employ the standard heatmap estimation loss~\cite{sun2019deep, liu2022temporal} $\mathcal{L}_H$ to optimize the GLSMamba framework.:
\begin{equation}
	\begin{aligned}
		\mathcal{L}_H = \left\|\hat{\mb{H}}_t^i - {\mb{H}}_t^i \right\|_2^2,
	\end{aligned}
\end{equation}
where $\hat{\mb{H}}_t^i$ and ${\mb{H}}_t^i$ denote the predicted and corresponding ground truth heatmaps, respectively.

\section{Experiments}
\label{sec:experiments}
\subsection{Experimental Settings} 
 
  \renewcommand\arraystretch{1.1}
\begin{table}[t]
\centering
  \resizebox{.49\textwidth}{!}{
  \begin{tabular}{l|l|ccccccc|c}
    \hline
      Method   &Backbone                         &Head   &Shoulder &Elbow       &Wrist   &Hip    &Knee   &Ankle   &{\bf Mean}\cr
      \hline
     PoseFlow~\cite{xiu2018pose}      &ResNet-152   &$66.7$ & $73.3$  &$68.3$      &$61.1$  &$67.5$ &$67.0$ &$61.3$  &{$ 66.5$}\cr
   FastPose~\cite{zhang2019fastpose}  &ResNet-101	&$80.0$ &$80.3$   &$69.5$      &$59.1$  &$71.4$ &$67.5$ &$59.4$  &{$ 70.3$}\cr
SimplePose~\cite{xiao2018simple}  &ResNet-152  &$81.7$ &$83.4$   &$80.0$      &$72.4$  &$75.3$ &$74.8$ &$67.1$  &{$ 76.7$}\cr
  STEmbedding~\cite{jin2019multi}     &Hourglass   &$83.8$ &$81.6$   &$77.1$      &$70.0$  &$77.4$ &$74.5$ &$70.8$  &{$ 77.0$}\cr
        HRNet~\cite{sun2019deep}      &HRNet-W48   &$82.1$ &$83.6$   &$80.4$      &$73.3$  &$75.5$ &$75.3$ &$68.5$  &{$ 77.3$}\cr
         MDPN~\cite{guo2018multi}     &ResNet-152   &$85.2$ &$88.5$   &$83.9$      &$77.5$  & $79.0$&$77.0$ &$71.4$  &{$ 80.7$}\cr
   CorrTrack~\cite{rafi2020self}   	  &CPN	&$86.1$ &$87.0$   &$83.4$      &$76.4$  & $77.3$&$79.2$ &$73.3$  &{$ 80.8$}\cr 
   Dynamic~\cite{yang2021learning}&HRNet-W48 	 &$88.4$ &$88.4$   &$82.0$      &$ 74.5$ &$79.1$ &$78.3$ &$73.1$  &{$81.1$}\cr
   PoseWarper~\cite{bertasius2019learning} &HRNet-W48 &$81.4$ &$88.3$   &$83.9$      &$ 78.0$ &$82.4$ &$80.5$ &$73.6$  &$81.2$\cr
   DCPose~\cite{liu2021deep} &HRNet-W48 &$ 88.0$  &$ 88.7$     &$ 84.1$   &$78.4$&$ 83.0$        &$ 81.4$&$ 74.2$ &$ 82.8$\cr
   DetTrack~\cite{wang2020combining} &HRNet-W48  &$89.4$       &$89.7$     &$85.5$ &$79.5$ &$82.4$      &$80.8$       &$76.4$   &$83.8$\cr
    FAMI-Pose~\cite{liu2022temporal} &HRNet-W48	&$ 89.6$  &$ 90.1$ &$ 86.3$ &$80.0$ &$ 84.6$ &$83.4$ &$ 77.0$ &$ 84.8$\cr
    TDMI~\cite{feng2023mutual} &HRNet-W48	&$ 90.0$  &$ 91.1$ &$ 87.1$ &$ 81.4$ &$ 85.2$ &$  84.5$ &$ 78.5$ &$ 85.7$\cr
   
    \hline  
	DiffPose~\cite{feng2023diffpose} & ViT-B &$ 89.0$  &$ 91.2$ &$ 87.4$ &$ 83.5$ &$ 85.5$ &$  87.2$ &$ 80.2$ &$ 86.4$\cr
	   DSTA~\cite{he2024video} & ViT-H &$ 89.3$  &$ 90.6$ &$ 87.3$ &$ 82.6$ &$ 84.5$ &$  85.1$ &$ 77.8$ &$ 85.6$\cr  
       \hline           
       \bf GLSMamba-B  & ViT-B			&$ 90.6$  	&$91.3$ &$  88.2$ &$ 83.8$ &$ 85.4$ &$ 87.1$ &$ 80.5$ &$\bf 86.9$\cr 
       \bf GLSMamba-H   & ViT-H		&$ 90.7$ &$ 92.1$ &$ 89.2$ &$ 85.3$ &$ 87.0$ &$  88.4$ &$ 82.4$ &$\bf 88.0$\cr 
    \hline
    \end{tabular}}
      \caption{{Quantitative results} on the \textbf{PoseTrack2017} validation set.}  \label{17val}
      \vspace{-.6em}
    \end{table}

 \renewcommand\arraystretch{1.1}
\begin{table}[t]
\centering
   \resizebox{0.49\textwidth}{!}{
   \begin{tabular}{l|l|ccccccc|c}
     \hline
      Method       &Backbone                      &Head &Shoulder &Elbow  &Wrist &Hip &Knee &Ankle &{\bf Mean}\cr
     \hline
 AlphaPose~\cite{fang2017rmpe}  &Hourglass         &$63.9$  &$78.7$&$77.4$ &$71.0$ &$73.7$ &$73.0$    &$69.7$     &{$71.9$}\cr
 MDPN~\cite{guo2018multi}       &ResNet-152         &$75.4$ &$81.2$ &$79.0$ &$74.1$ &$72.4$ &$73.0$  &$69.9$   &{$75.0$}\cr
 PGPT~\cite{bao2020pose}    	&ResNet-152 		&-       &-     &-      &$72.3$ &-      &-       &$72.2$   &{$76.8$}\cr
 Dynamic~\cite{yang2021learning} &HRNet-W48 	&$80.6$ &$84.5$   &$80.6$  &$ 74.4$ &$75.0$ &$76.7$ &$71.8$  &$77.9$\cr
 PoseWarper~\cite{bertasius2019learning}&HRNet-W48  &$79.9$&$86.3$&$82.4$&$77.5$&$79.8$&$78.8$&$73.2$  &$79.7$\cr
 PT-CPN++~\cite{yu2018multi}	&CPN  &$82.4$ &$88.8$ &$86.2$ &$79.4$ &$72.0$ &$80.6$ &$76.2$  &$80.9$\cr
 DCPose~\cite{liu2021deep} 		&HRNet-W48 &$ 84.0$ &$ 86.6$&$ 82.7$&$ 78.0$&$ 80.4$&$ 79.3$&$ 73.8$&$ 80.9$\cr 
 DetTrack~\cite{wang2020combining} &HRNet-W48 &$84.9$ &$87.4$ &$84.8$ &$79.2$ &$77.6$      &$79.7$       &$75.3$   &$81.5$ \cr
 FAMI-Pose~\cite{liu2022temporal}&HRNet-W48 &$ 85.5$&$ 87.7$&$ 84.2$&$ 79.2$&$ 81.4$&$81.1$&$ 74.9$&$ 82.2$\cr
 TDMI~\cite{feng2023mutual}&HRNet-W48	 &$ 86.2$&$ 88.7$&$ 85.4$&$ 80.6$&$ 82.4$&$ 82.1$&$ 77.5$&$ 83.5$\cr
 \hline
 DiffPose~\cite{feng2023diffpose} &ViT-B	 &$ 85.0$&$ 87.7$&$ 84.3$&$ 81.5$&$ 81.4$&$ 82.9$&$ 77.6$&$ 83.0$\cr
 DSTA~\cite{he2024video} & ViT-H &$ 85.9$  &$ 88.8$ &$ 85.0$ &$ 81.1$ &$ 81.5$ &$  83.0$ &$ 77.4$ &$ 83.4$\cr 
  \hline  
      \bf GLSMamba-B  & ViT-B	&$ 85.0$  &$88.2$ &$  85.6$ &$ 82.9$ &$ 82.5$ &$ 84.9$ &$ 79.7$ &$\bf 84.2$\cr 
   \bf GLSMamba-H  & ViT-H	&$ 85.6$  &$ 88.9$ &$ 86.5$ &$83.6$ &$ 82.9$ &$ 85.7$ &$81.4$ &$\bf 84.9$\cr 
     \hline
     \end{tabular}}
     \caption{{Quantitative results} on the \textbf{PoseTrack2018} validation set.}  \label{18val}
     \vspace{-1.2em}
   \end{table}

 \renewcommand\arraystretch{1.1}
\begin{table}[t]
\centering
   \resizebox{0.49\textwidth}{!}{
   \begin{tabular}{l|l|ccccccc|c}
     \hline
     Method  &Backbone  &Head &Shoulder &Elbow  &Wrist &Hip &Knee &Ankle &{\bf Mean}\cr
     \hline
Simple~\cite{xiao2018simple} 	 &ResNet-152  &$80.5$ &$81.2$ &$73.2$ &$64.8$ &$73.9$ &$72.7$  &$67.7$   &$73.9$\cr
HRNet~\cite{sun2019deep}           	 &HRNet-W48   &$81.5$ &$83.2$ &$81.1$ &$75.4$ &$79.2$ &$77.8$  &$71.9$   &$78.8$\cr
 PoseWarper~\cite{bertasius2019learning} &HRNet-W48   &$82.3$ &$84.0$ &$82.2$ &$75.5$ &$80.7$ &$78.7$  &$71.6$   &$79.5$\cr
DCPose~\cite{liu2021deep}				 &HRNet-W48   &$ 83.2$&$ 84.7$&$ 82.3$&$ 78.1$&$ 80.3$&$ 79.2$&$ 73.5$&$ 80.5$\cr 
 FAMI-Pose~\cite{liu2022temporal} 		 &HRNet-W48   &$ 83.3$&$ 85.4$&$ 82.9$&$ 78.6$&$ 81.3$&$80.5$&$ 75.3$&$ 81.2$\cr
 TDMI~\cite{feng2023mutual} &HRNet-W48	  &$ 85.8$&$ 87.5$&$ 85.1$&$ 81.2$&$ 83.5$&$82.4$&$ 77.9$&$ 83.5$\cr
 \hline
    DiffPose~\cite{feng2023diffpose}		 &ViT-B	  &$ 84.7$&$ 85.6$&$ 83.6$&$ 80.8$&$ 81.4$&$83.5$&$ 80.0$&$ 82.9$\cr
 DSTA~\cite{he2024video} 				 &ViT-H 	  &$ 87.5$&$ 87.0$ &$ 84.2$ &$ 81.4$ &$ 82.3$ &$  82.5$ &$ 77.7$ &$ 83.5$\cr 
   \hline 
 \bf GLSMamba-B  & ViT-B	&$ 86.3$  &$86.7$ &$  85.1$ &$ 82.1$ &$ 83.0$ &$ 84.3$ &$ 79.4$ &$\bf 84.1$\cr 
\bf GLSMamba-H  & ViT-H	&$ 87.0$  &$ 86.9$ &$  85.4$ &$ 83.2$ &$ 83.4$ &$  84.8$ &$ 80.8$ &$\bf 84.7$\cr 
     \hline 
     \end{tabular}}
     \caption{{Quantitative results} on the \textbf{PoseTrack21} dataset. } \label{21val}
    \vspace{-.3em}
   \end{table} 
 
 \renewcommand\arraystretch{1.1}
\begin{table}
  \resizebox{0.49\textwidth}{!}{
  \begin{tabular}{l|l|ccccccc|c}
    \hline
     Method     &Backbone                       &Head&Shoulder &Elbow  &Wrist &Hip &Knee &Ankle &{\bf Avg}\cr
    \hline
Thin-slicing~\cite{song2017thin} &$-$   &$97.1$      &$95.7$     &$87.5$      &$81.6$ &$98.0$   &$92.7$   &$89.8$   &{$92.1$}\cr
LSTM PM~\cite{luo2018lstm}    	&$-$	&$98.2$ &$96.5$ &$89.6$ &$86.0$ &$98.7$ &$95.6$  &$90.0$   &{$93.6$}\cr
DKD~\cite{nie2019dynamic} &ResNet-50    &$98.3$ &$96.6$ &$90.4$ &$87.1$ &$ 99.1$ &$ 96.0$  &$92.9$   &{$94.0$}\cr
K-FPN~\cite{zhang2020key} &ResNet-18	&$94.7$ &$96.3$ &$ 95.2$ &$90.2$ &$96.4$ &$95.5$  &$93.2$   &{$94.5$}\cr
K-FPN~\cite{zhang2020key} &ResNet-50	&$95.1$ &$96.4$ &$ 95.3$ &$91.3$ &$96.3$ &$95.6$  &$92.6$   &{$94.7$}\cr 
MAPN~\cite{fan2021motion} &ResNet-18	&$98.2$ &$97.4$ &$ 91.7$ &$85.2$ &$ 99.2$ &$ 96.7$  &$92.2$   &{$94.7$}\cr 
FAMI-Pose~\cite{liu2022temporal} &HRNet-W48 &$ 99.3$&$ 98.6$&$ 94.5$&$ 91.7$&$ 99.2$&$91.8$&$ 95.4$&$ 96.0$\cr
\hline
DeciWatch~\cite{zeng2022deciwatch}\ddag  &SimplePose     &$ 99.8$ &$99.5$ &$ 99.7$ &$ 99.7$ &$ 98.7$ &$ 99.4$  &$96.5$   &{$\bf 98.8$}\cr 
\hline
  \bf GLSMamba-B  & ViT-B	&$ 99.2$  &$ 98.3$ &$ 98.1$ &$97.1$ &$99.3$ &$ 98.0$ &$ 95.9$ &$ 97.9$\cr 
    \hline
    \end{tabular}}
    \caption{Quantitative results on the \textbf{Sub-JHMDB} dataset. 
    } \label{jhmdb}
    \vspace{-1.4em}
\end{table}

\begin{figure*}
\begin{center}
\includegraphics[width=.8\linewidth]{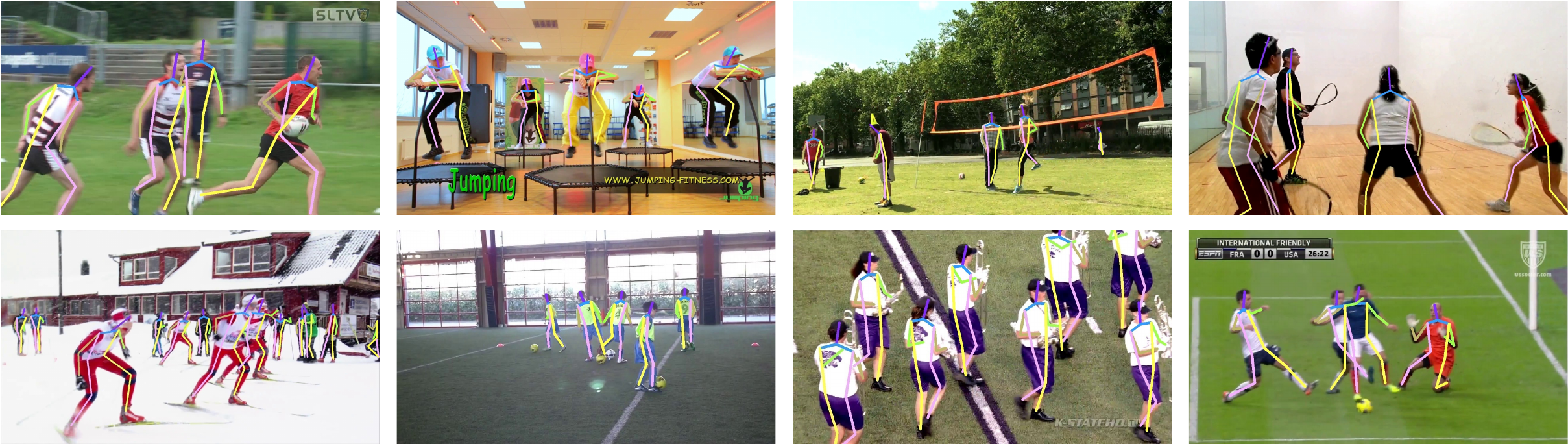}
\end{center}
\vspace{-1em}
\caption{Visual results of our method on benchmarks. Challenging scenes such as occlusion and motion blur are involved.}\vspace{-1.2em}
\label{fig:vis}
\end{figure*}

\textbf{Datasets and evaluation.}\quad
 We evaluate our approach on four challenging VHPE benchmarks, including PoseTrack2017~\cite{Iqbal_2017_CVPR}, PoseTrack2018~\cite{Andriluka_2018_CVPR}, PoseTrack21~\cite{doering2022posetrack21}, and Sub-JHMDB~\cite{jhuang2013towards}. Specifically, \textbf{PoseTrack2017} provides $80,144$ human pose annotations which are divided into \emph{train}/\emph{val} sets, consisting of $250$  and $50$ video clips, respectively. \textbf{PoseTrack2018} significantly expands the amount of data and contains $593$ videos for training and $170$ for validation, with a total of $153,615$ manually labeled poses. Both datasets are annotated with $15$ keypoints along the same criteria, and include an extra flag for visibility. \textbf{PoseTrack21} further enriches the annotations of {PoseTrack2018} especially for complex small persons, providing $177,164$ human pose labels. 
 \textbf{Sub-JHMDB} contains 316 video sequences with $11,200$ frames. Following \cite{liu2022temporal}, we adopt three data splits for training and testing, and report the average performance. We benchmark the model over visible joints using the metric of average precision (AP)~\cite{sun2019deep, liu2021deep}. 

\noindent
 \textbf{Implementation details.}\quad The proposed GLSMamba framework is implemented by PyTorch. We incorporate data augmentations such as random rotation/scaling, truncation, and flipping in the training phase. We take ViTPose pretrained on COCO as the backbone, and freeze its parameters during training. The temporal span $\delta$ is set to $2$.  We employ AdamW optimizer with a base learning rate of $1e-4$, which decays to $1e-5$ at $6$-th epoch and $1e-6$ for $12$-th epoch. All training process is performed on one TITAN RTX GPU and terminated within $20$ epochs.

 \subsection{Comparison with State-of-the-art Approaches}
 We first compare GLSMamba with state-of-the-art (SOTA) methods on the PoseTrack2017 validation set, and report the results in Table \ref{17val}. We comprehensively evaluate GLSMamba under two widely-used backbones namely ViT-B and ViT-H, and provide the computational cost in Table \ref{run}. We observe that GLSMamba, the first pure Mamba-based VHPE framework with only $9.8$ M trainable parameters, delivers SOTA pose estimation performance against existing well-established CNN- and Transformer-based models across various backbones. \textbf{(i)} Compared to the impressive convolution-based PoseWarper~\cite{bertasius2019learning}, GLSMamba-B attains a remarkable performance gain of $5.7$ mAP with drastically reduced FLOPs ($\downarrow 34\%$). GLSMamba-B also improves the pose estimation performance by $1.2$ mAP over the SOTA method TDMI~\cite{feng2023mutual}. Compared to the Transformer-based DiffPose~\cite{feng2023diffpose} that operates on low-resolution sequences, GLSMamba-B improves the mAP by $0.5$ points. Such compelling results demonstrate the importance of explicitly embracing both global and local high-resolution spatiotemporal contexts, reflecting the great potential of the novel Mamba-based architecture for VHPE.
 Noticeably, in contrast to existing SOTA methods~\cite{bertasius2019learning, liu2021deep, feng2023diffpose} that often additionally fine-tune the backbone on VHPE datasets to improve performance, we directly leverage the pre-trained backbone weights on COCO from \cite{xu2022vitpose}. This simplifies the training pipeline, and remarkably diminishes the trainable parameters $\dag$ by $\downarrow 86.2\%$. \textbf{(ii)} When adopting the larger backbone ViT-H, GLSMamba-H further pushes forward the performance boundary and achieves $88.0$ mAP ($\uparrow 1.6$).

 Table \ref{18val} and Table \ref{21val} provide the experimental comparisons of various approaches on the PoseTrack2018 and PoseTrack21 datasets, respectively. With the base backbone ViT-B, our GLSMamba-B has already surpassed all other methods in both datasets. Our large model, GLSMamba-H, further obtains new state-of-the-art performance of $84.9$ mAP and $84.7$ mAP. We also illustrate in Fig.~\ref{fig:vis} the example visualizations of pose estimates in complex scenarios, which attest to the effectiveness of the proposed method.
 
Furthermore, we benchmark the proposed model on Sub-JHMDB and tabulate the results in Table \ref{jhmdb}. Compared to the SOTA representation learning approach FAMI-Pose \cite{liu2022temporal}, GLSMamba-B can provide a significant performance improvement of $1.9$ mAP. On the other hand, in contrast to the best-performed post-processing method \cite{zeng2022deciwatch} $\ddag$ that operates in the pose coordinate space, our GLSMamba-B still achieves a competitive performance of $97.9$ mAP. 

\begin{figure}[t]
\begin{center}
\includegraphics[width=.95\linewidth]{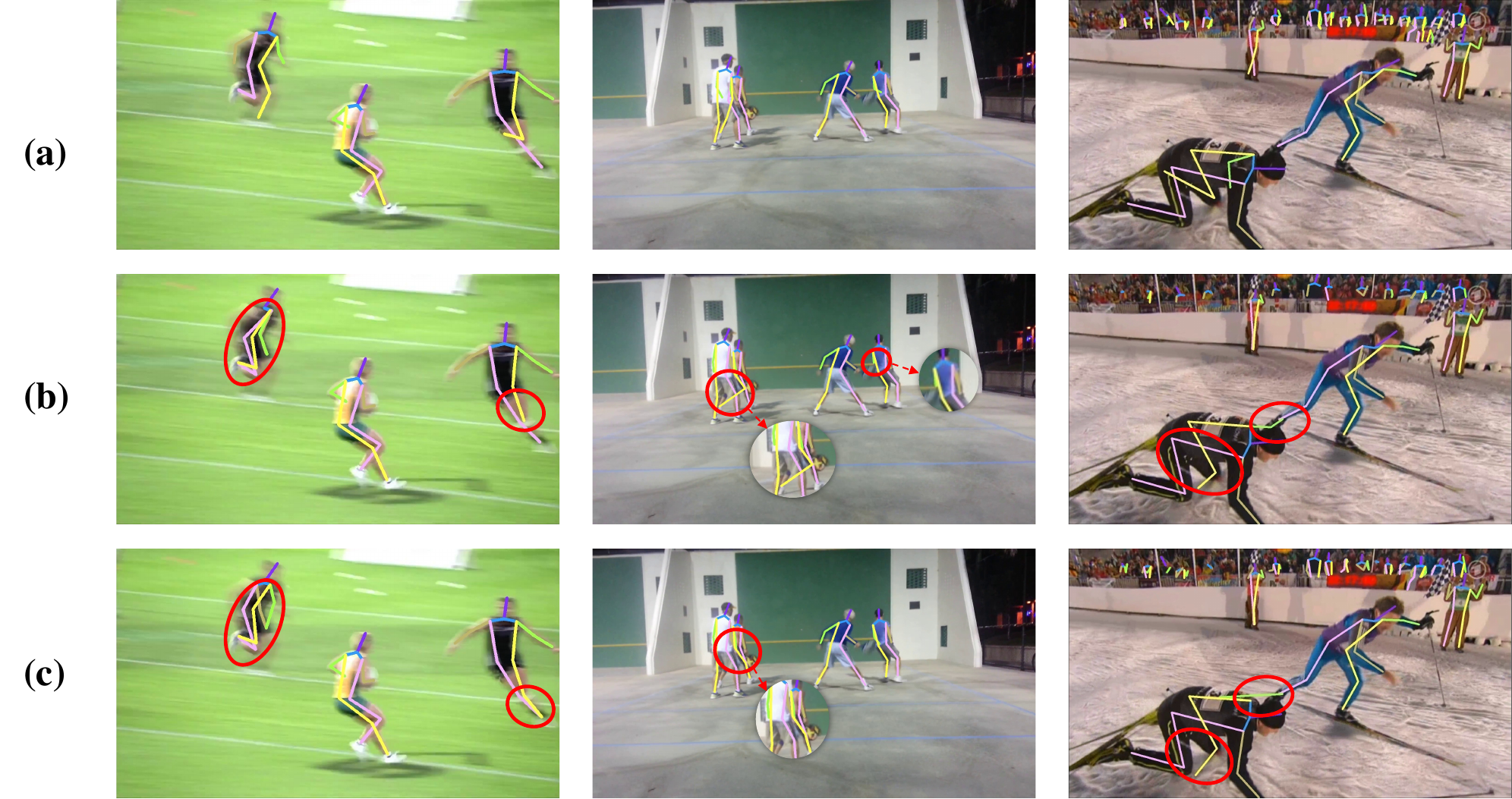}
\end{center}
\vspace{-1.em}
\caption{Qualitative comparisons of pose predictions of (a) GLSMamba-B, (b) TDMI, and (c) DiffPose on the PoseTrack dataset. Inaccurate results are highlighted by red circles.}  
\vspace{-1.5em}
\label{fig:com}
\end{figure}

\noindent
\textbf{Qualitative analyses.}\quad In addition to the quantitative comparisons, we also qualitatively examine the ability of GLSMamba to cope with challenging scenes. As illustrated in Fig.~\ref{fig:com}, we present the side-by-side comparisons of the proposed method (a) against SOTA models TDMI~\cite{feng2023mutual} (b) and DiffPose~\cite{feng2023diffpose} (c). Remarkably, our approach achieves more robust and accurate results across various scenarios. TDMI is built upon convolutions that suffer limited receptive fields, leading to suboptimal performance. On the other hand, DiffPose leverages self-attentions and overlooks rich keypoint motional details. Through the principled design of GSM and LRM, our method can capture reliable global-local high-resolution spatiotemporal representations and is more adept at handling complex cases.

  \renewcommand\arraystretch{1.}
\begin{table}[t]
\centering
   \resizebox{0.44\textwidth}{!}{
   \begin{tabular}{l|cc|c}
     \hline
     Method &Global Spat. Mamba (GSM) &Local Ref. Mamba (LRM) & mAP\cr
     \hline
	(a) Backbone & &  &$74.2$\cr
	\hline
	(b) GSM &$\checkmark$ &  &$86.0$\cr
	(c) GLSMamba-B  &$\checkmark$ &$\checkmark$  &$\bf 86.9$\cr
      \hline
     \end{tabular}}
     \vspace{-.3em}
     \caption{Ablation study of different components.} \label{abl:com}
     \vspace{-.6em}
   \end{table}   
   
 \renewcommand\arraystretch{1.}
\begin{table}[t]
\centering
   \resizebox{0.44\textwidth}{!}{
   \begin{tabular}{l|c|c|c}
     \hline
     Methods &\#Params. &GFLOPs & mAP\cr
     \hline
	(a) unified scanning &$9.1$ M &$137.4$ &$85.8$\cr
	(b) unified + space-wise scanning &$9.4$ M &$138.1$ &$86.5$\cr
    (c) unified + space-wise + time-wise scanning &$9.8$ M &$138.9$ &$\bf 86.9$\cr
      \hline
   (d) w/o STMM    &$9.1$ M &$137.4$ &$86.2$\cr
   \hline
     \end{tabular}}
     \vspace{-.3em}
     \caption{Ablation study of STS6D and STMM.}\vspace{-1.7em} \label{abl:scan}
   \end{table}   

 \subsection{Ablation Study}
In this section, we investigate the impact of each proposed component and  design choice in GLSMamba-B. All experiments are performed on the PoseTrack2017 validation set.

\noindent
 \textbf{Study on components.}\quad 
We first study the contribution of each individual component including Global Spatiotemporal Mamba (GSM) and Local Refinement Mamba (LRM), and provide the empirical results in Table \ref{abl:com}. \textbf{(a)} For the first setting, we remove both proposed GSM and LRM modules, and estimate human poses employing only the backbone (ViT-B). This baseline obtains a $74.2$ mAP. \textbf{(b)} Subsequently, we incorporate the GSM module on top of the backbone (a) for global dynamic modeling, which significantly improves upon the baseline by a large margin of $11.8$ mAP and is on par with the SOTA approach DiffPose \cite{feng2023diffpose}. This corroborates the effectiveness of our method in introducing global spatiotemporal knowledge to facilitate VHPE. \textbf{(c)} For the final setting, we further introduce LRM which corresponds to the complete GLSMamba-B model. The performance improvement of $0.9$ mAP suggests the importance of capturing enriched high-frequency details of local keypoint motions for accurate pose estimation.

\noindent
 \textbf{Study on GSM designs.}\quad
 Then, we validate the efficacy of the core GSM designs, including the 6D selective Space-Time Scan (STS6D) and Spatial- and Temporal-Modulated scan Merging (STMM). 
   \textbf{(1)} As presented in Table \ref{abl:scan} (a)-(c), we gradually introduce diverse scanning directions containing unified, space-wise, and time-wise scanning routes. The results in mAP reflect a progressive and remarkable performance improvement, from $85.8 \rightarrow 86.5 \rightarrow 86.9$, with negligible extra computations. This is in line with our expectations, \emph{i.e.}, an adequate space-time traversal allows for effective mining of dense high-resolution sequence knowledge, thereby contributing to enhanced accuracy. \textbf{(2)} We also examine the impact of the proposed STMM strategy by removing it and merging diverse scans via a simple addition. The significant performance reduction of $0.7$ mAP (d) highlights the importance of STMM in adaptively aggregating distinct scanning knowledge.
 
 \noindent
 \textbf{Comparison with VideoMamba}.\quad We notice that VideoMamba~\cite{li2024videomamba, park2024videomamba} has proposed the latest Mamba-based framework for high-level video understanding. However, our method differs notably from VideoMamba: (1) For the global modeling, we introduce a Sequential Channel Attention to filter unnecessary information, and design STS6D and STMM for adequate spatiotemporal scanning and adaptive fusion.  (2) Unlike VideoMamba that lacks of the local modeling capability, we also propose a Windowed Space-Time Scan (WSTS) to enhance local details. 

\noindent 
 \textbf{Spatiotemporal representation resolution.}\quad
Finally, we examine the influence of feature resolutions on the pose estimation performance. As reported in Table \ref{run}, the following baselines are constructed: \textbf{a)} We directly adapt GLSMamba-B to low-resolution sequences ($\frac{1}{16}\mathtt{H} \times \frac{1}{16}\mathtt{W} \times \mathtt{T}$) which forms \emph{GLSMamba-BLR*}. \textbf{b)} We employ ViT-B as backbone and stack six standard ViT-B blocks to learn spatiotemporal features at low (\emph{TransLR*}), normal (\emph{TransNR*}), and high (\emph{TransHR*}) resolutions, respectively. It is observed that spatiotemporal representations with higher resolutions indeed result in better performance, across both Mamba ($\uparrow 1.2$ mAP) and Transformer ($\uparrow 0.6$ mAP) architectures.
 This is in line with our intuitions that high-resolution sequence representations can capture inter-frame temporal dynamics and intra-frame spatial details more precisely, which facilitate accurate pose heatmaps. Another observation is that high-resolution settings lead to significantly increased computational overhead, especially for Transformer structures (Out Of Memory (OOM) vs $138.9$G FLOPs at $15,360$ input tokens).
  This highlights that Mamba can achieve better computational trade-offs in handling high-resolution feature sequences.

\renewcommand\arraystretch{1.2}
\begin{table}[t]
\centering
   \resizebox{0.49\textwidth}{!}{
   \begin{tabular}{l|lc|cc|c}
     \hline
     Method  &Resolution &Token Num. &\#Params.  &GFLOPs & Mean\cr
     \hline
   GLSMamba-B   &$1/4 \times T$ & $15,360$  &$\bf 9.8$ M\dag &$\bf 138.9$ &$ 86.9 (\uparrow 1.2)$\cr
  GLSMamba-BLR*   &$1/16 \times T$ & $960$  &$ 9.8$ M\dag &$ 85.1$ &$ 85.7$\cr 
 \hline
 TransLR*		&$1/16 \times T$ & $960$ &$46.3$ M\dag &$125.7$ &$ 84.2$\cr
 TransNR*		&$1/8 \times T$ & $3,840$ &$47$ M\dag &$315.2$ &$ 84.8 (\uparrow 0.6)$\cr
 TransHR*		&$1/4 \times T$ & $15,360$ &$-$ &$-$ &\textbf{OOM} \cr
 \hline
    PoseWarper~\cite{bertasius2019learning}		&$1/4 \times T$ & $-$ &$71.1$ M\dag &$210.5$ &$ 81.2$\cr
      \hline
     \end{tabular}}
     \vspace{-.2em}
     \caption{Impact of feature sequence resolutions. ``\dag" denotes trainable parameters and ``*" indicates manually-constructed baselines.} \label{run}
     \vspace{-1.4em}
   \end{table}

\vspace{-.5em}
\section{Conclusion and Future Works} 
This paper introduces GLSMamba, a novel framework that leverages State Space Models to learn decoupled global and local high-resolution spatiotemporal representations for VHPE. We design a Global Spatiotemporal Mamba with 6D selective space-time scan and spatial- and temporal-modulated scan merging mechanisms, to fully analyze holistic human dynamics embedded in dense high-resolution spatiotemporal contexts from a global perspective. A Local Refinement Mamba based on windowed space-time scan is further introduced for enhancing localized keypoint motion details. Extensive experiments on four benchmarks demonstrate the superiority of GLSMamba in both performance and computational trade-offs.
Future works include applications for other vision tasks such as 3D human pose estimation and video segmentation.

\vspace{-.6em}
\section{Acknowledgements}
This research was supported by the National Natural Science Foundation of China under Grant Nos. 62203184 and W2421093, and the International Cooperation Project of Jilin Province under Grant No.~20250205079GH. 
This research was also supported by the National Key R\&D Program of China under Grant No. 2023YFF0905400 and the National Natural Science Foundation of China through Grant No.~U2341229.
This research was also supported by the MSIT (Ministry of Science and ICT), Korea, under the ITRC (Information Technology Research Center) support program (IITP-2025-RS-2020-II201789) supervised by the IITP (Institute for Information \& Communications Technology Planning \& Evaluation). 

{
    \small
    \bibliographystyle{ieeenat_fullname}
    \bibliography{References}
}

\section{Appendix}

\subsection{Differences against Mamba-Based Methods}
We notice that several works~\cite{huang2025posemamba, zhang2025pose, dang2024mamkpd, zhang2024mambapose, yang2024vimpose} have applied Mamba to human pose estimation-related tasks. Compared to these approaches, our distinct contributions are summarized as:
\begin{enumerate}
	\item  While existing methods fall within 3D/Multi-Person Pose Estimation and operate on 2D skeleton sequences or single images, we present \emph{the first Mamba-based video pose estimation (VHPE) model capable of processing more challenging video sequences with higher information density.}
	\item  Unlike most hybrid architectures (PoseMagic~\cite{zhang2025pose}, MambaPose~\cite{zhang2024mambapose}, MamKPD~\cite{dang2024mamkpd}, ViMPose~\cite{yang2024vimpose}) combining Mamba with GCNs/CNNs, we design a \emph{pure Mamba framework for both global and local modeling.} In contrast to PoseMamba~\cite{huang2025posemamba} that employs local limb scanning to capture skeleton spatial dependencies, we devise \emph{a windowed space-time scan to enhance local keypoint motion details.}
	\item The core Mamba operator of existing methods performs bidirectional scanning in space/time domains~\cite{liu2024vmambavisualstatespace, li2024videomamba}, and simply sums different scanning results. Instead, \emph{we propose STS6D to fully resolve feature sequences from six directions, and STMM to adaptively aggregate diverse scanning knowledge.} 
\end{enumerate}

\begin{figure*}
\begin{center}
\includegraphics[width=.98\linewidth]{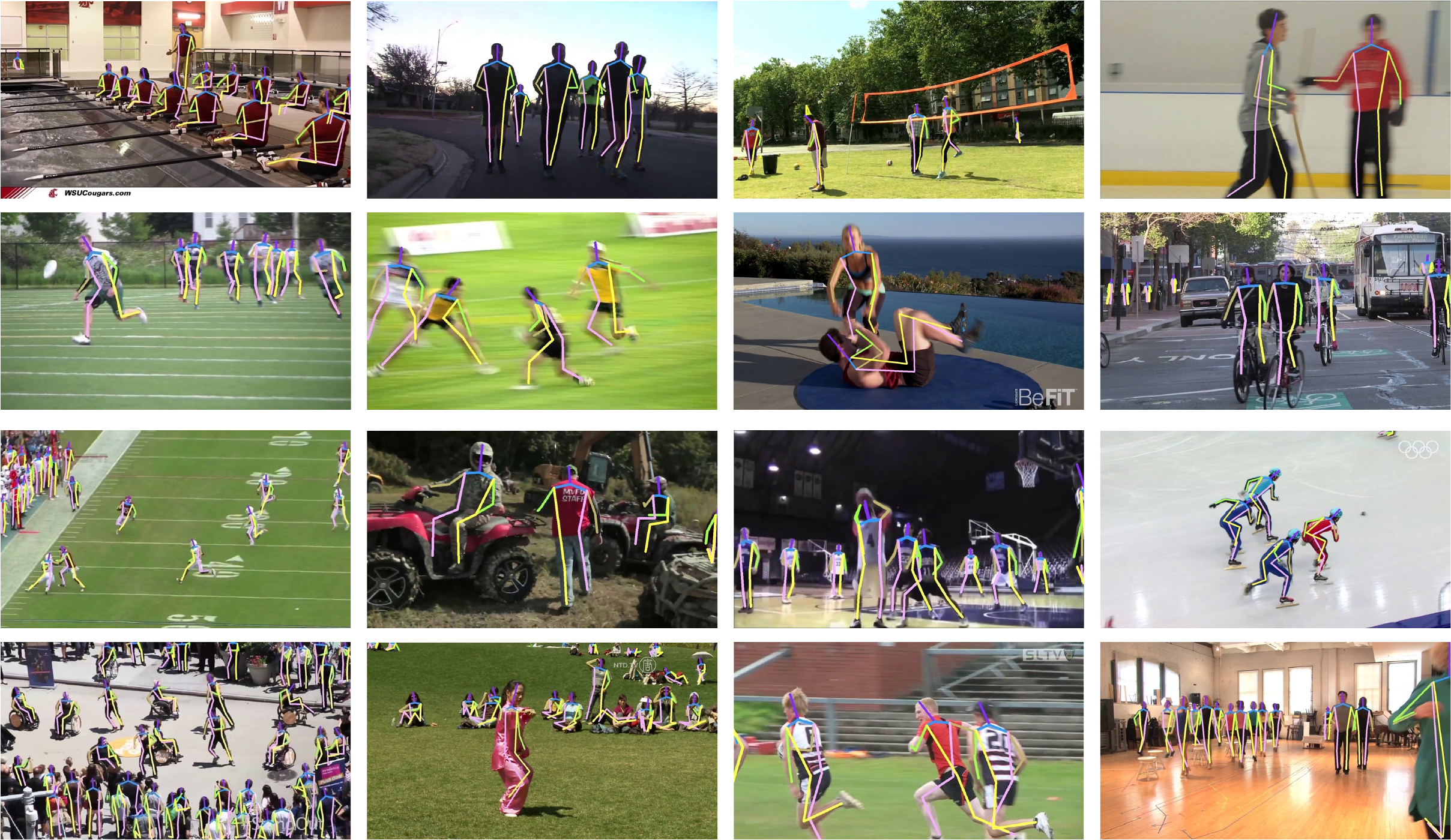}
\end{center}
\vspace{-1em}
\caption{Visual results of our method on the PoseTrack2017 dataset. Challenging scenes such as occlusion and motion blur are involved.}
\label{fig:17}
\end{figure*}

\begin{figure*}[!h]
\begin{center}
\includegraphics[width=.98\linewidth]{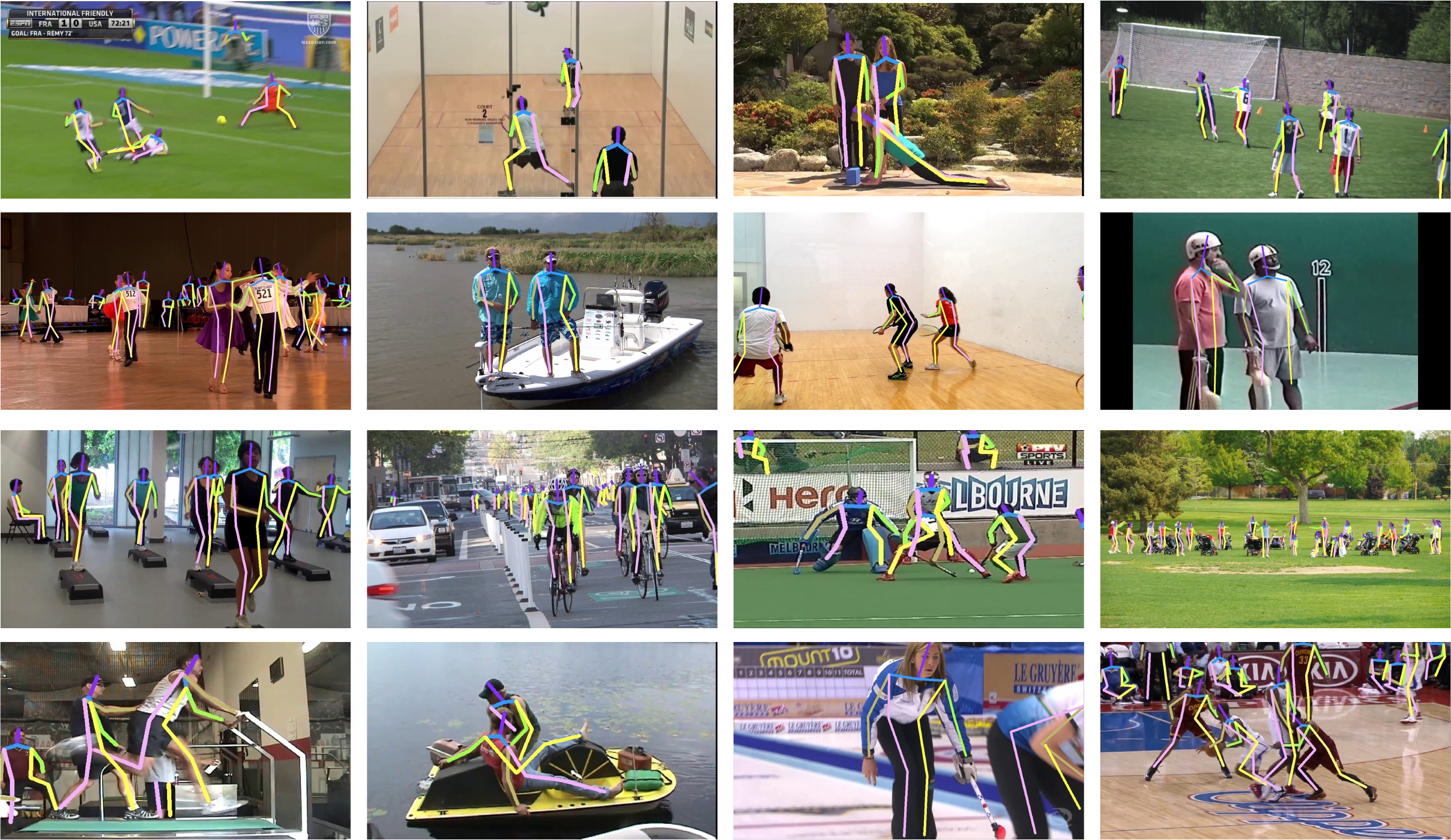}
\end{center}
\vspace{-1em}
\caption{Visual results of our method on the PoseTrack2018 dataset. Challenging scenes such as occlusion and motion blur are involved.}
\label{fig:18}
\end{figure*}

\begin{figure*}
\begin{center}
\includegraphics[width=.95\linewidth]{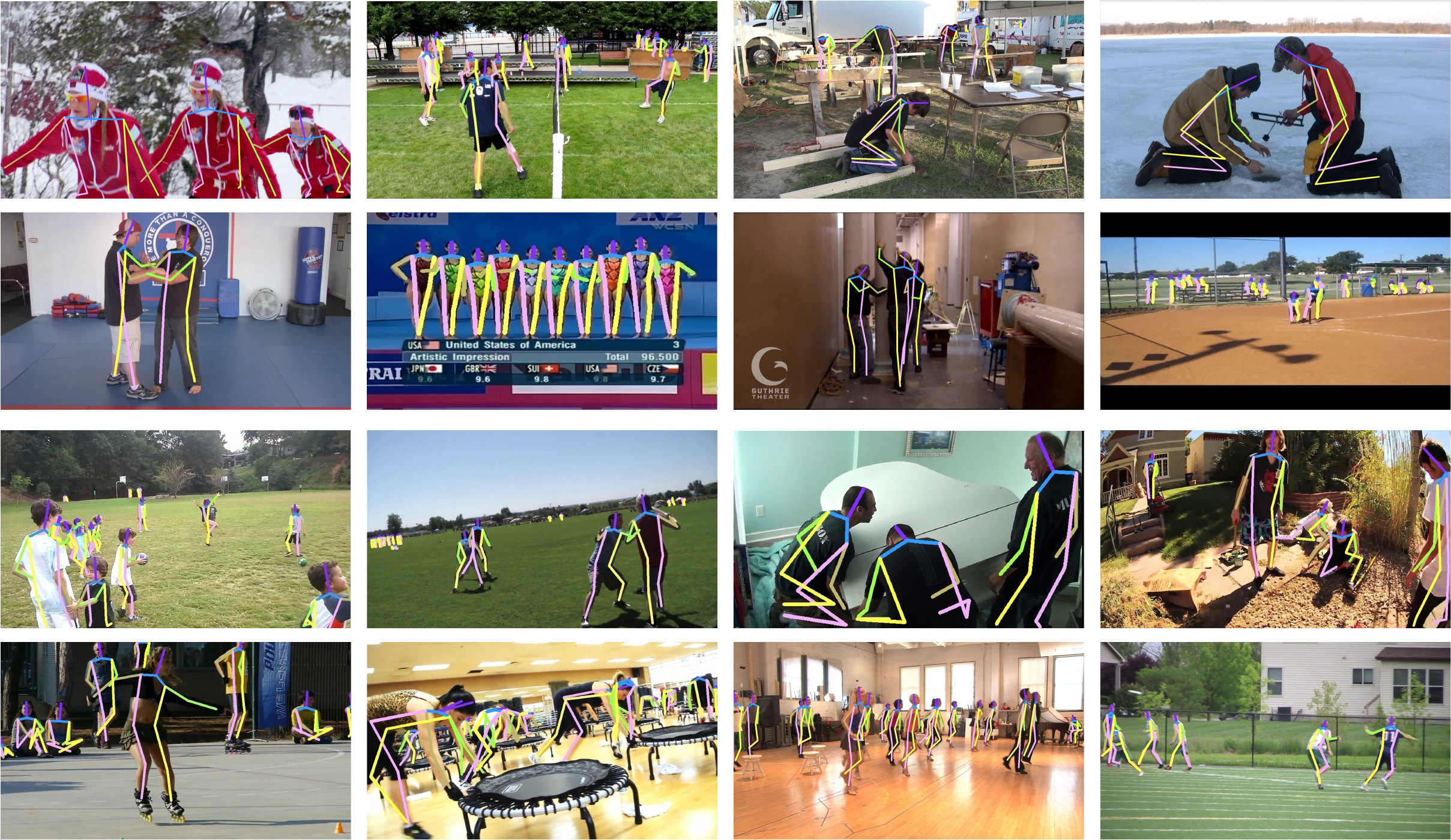}
\end{center}
\caption{Visual results of our method on the PoseTrack21 dataset. Challenging scenes such as occlusion and motion blur are involved.}
\label{fig:21}
\end{figure*}

\begin{figure*}[!h]
\begin{center}
\includegraphics[width=.98\linewidth]{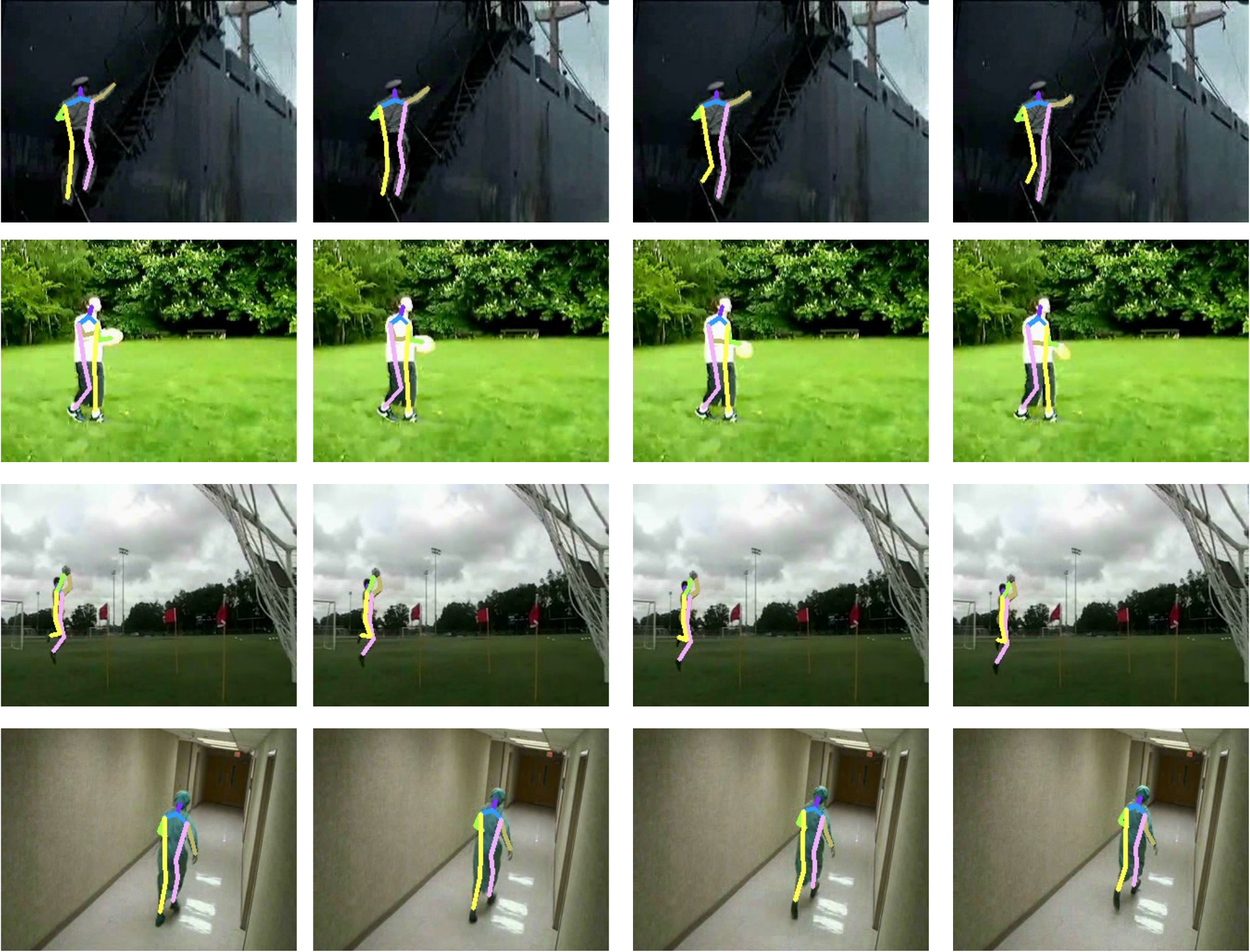}
\end{center}
\caption{Visual results of our method on the Sub-JHMDB dataset. Challenging scenes such as occlusion and motion blur are involved.}
\label{fig:jhmdb}
\end{figure*}

\subsection{Additional Qualitative Examples}
In this section, we present more visualized results of our proposed method. Figs.~\ref{fig:17}--\ref{fig:jhmdb} display our pose estimation results in PoseTrack2017~\cite{Iqbal_2017_CVPR}, PoseTrack2018~\cite{Andriluka_2018_CVPR}, PoseTrack21~\cite{doering2022posetrack21}, and Sub-JHMDB~\cite{jhuang2013towards} datasets, respectively. From these figures, we can observe that our method consistently achieves accurate and robust pose estimations in challenging scenes including mutual occlusion and motion blur.

\end{document}